\documentclass{article} % For LaTeX2e
\usepackage{iclr2024_conference,times}

\usepackage{amsmath,amsfonts,bm}

\def\eqref#1{equation~\ref{#1}}
\def\1{\bm{1}}

\def\vg{{\bm{g}}}

\def\vx{{\bm{x}}}

\DeclareMathAlphabet{\mathsfit}{\encodingdefault}{\sfdefault}{m}{sl}
\SetMathAlphabet{\mathsfit}{bold}{\encodingdefault}{\sfdefault}{bx}{n}

\definecolor{citecolor}{HTML}{103C5B}
\definecolor{linkcolor}{rgb}{0.956,0.298,0.235} 
\usepackage[hidelinks,colorlinks,linkcolor=linkcolor,citecolor=citecolor]{hyperref}
\usepackage{url}
\usepackage{graphicx}
\usepackage{gensymb}
\usepackage{amssymb}
\usepackage{tabularx}
\usepackage{booktabs}
\usepackage{multirow}
\usepackage{blindtext}
\usepackage{wrapfig}
\usepackage{algorithm}
\usepackage{algorithmic}
\usepackage{amsthm}
\usepackage{amssymb}

\newcommand{\ie}{\textit{i.e.}}

\usepackage{caption}
\usepackage[percent]{overpic}

\title{DreamCraft3D: Hierarchical 3D Generation with Bootstrapped Diffusion Prior}

\author{Jingxiang Sun$^{1}$\footnotemark[1] , Bo Zhang$^{3}$\footnotemark[2] , Ruizhi Shao$^{1}$, Lizhen Wang$^{1}$, Wen Liu$^{2}$, Zhenda Xie$^{2}$, Yebin Liu$^{1}$\footnotemark[2] \\
$^{1}$ Tsinghua University, $^{2}$ DeepSeek AI, $^{3}$ Independent Researcher \\ 
}

{
  \renewcommand{\thefootnote}%
    {\fnsymbol{footnote}}
  \footnotetext[1]{Work partially done during the internship at DeepSeek AI.}
  \footnotetext[2]{Corresponding authors.}
}

\iclrfinalcopy % Uncomment for camera-ready version, but NOT for submission.
\begin{document}

\maketitle

\begin{abstract}
% In this work, we introduce a hierarchical text-to-3D pipeline: first generating images from text prompts and then elevating these images to 3D via a tiered coarse-to-fine optimization. Central to our approach is an iterative Dreambooth-mesh optimization strategy. This employs 3D renderings to assemble a pseudo multi-view dataset, refining a multi-view Dreambooth that, in turn, optimizes the 3D representation iteratively. As the process continues, our experiments reveal that the generated pseudo datasets become more consistent and realistic. Iteratively, the pseudo datasets show heightened consistency and realism. To bolster texture realism, we embed variational score distillation (VSD) and present enhancements to VSD, incorporating elements like structure-aware latent regularization and a masked dataset refresh. Our methodology produces 3D content that boasts consistent textures, intricate geometry, and high fidelity. Through both quantitative and qualitative assessments, our approach is shown to surpass current state-of-the-art techniques in rendering quality and consistency.
We present DreamCraft3D, a hierarchical 3D content generation method that produces high-fidelity and coherent 3D objects. We tackle the problem by leveraging a 2D reference image to guide the stages of geometry sculpting and texture boosting. A central focus of this work is to address the consistency issue that existing works encounter. To sculpt geometries that render coherently, we perform score distillation sampling via a view-dependent diffusion model. This 3D prior, alongside several training strategies, prioritizes the geometry consistency but compromises the texture fidelity. We further propose \textit{Bootstrapped Score Distillation} to specifically boost the texture. We train a personalized diffusion model, Dreambooth, on the augmented renderings of the scene, imbuing it with 3D knowledge of the scene being optimized. The score distillation from this 3D-aware diffusion prior provides view-consistent guidance for the scene. Notably, through an alternating optimization of the diffusion prior and 3D scene representation, we achieve mutually reinforcing improvements: the optimized 3D scene aids in training the scene-specific diffusion model, which offers increasingly view-consistent guidance for 3D optimization. The optimization is thus bootstrapped and leads to substantial texture boosting. With tailored 3D priors throughout the hierarchical generation, DreamCraft3D generates coherent 3D objects with photorealistic renderings, advancing the state-of-the-art in 3D content generation. Code available at \url{https://github.com/deepseek-ai/DreamCraft3D}.

\end{abstract}

\section{Introduction}

The remarkable success of 2D generative modeling~\citep{saharia2022photorealistic,ramesh2022hierarchical,rombach2022high,gu2022vector} has profoundly shaped the way that we create visual content. 3D content creation, which is crucial for applications like games, movies and virtual reality, still presents a significant challenge for deep generative networks. While 3D generative modeling has shown compelling results for certain categories~\citep{wang2023rodin,chan2022efficient,zhang2023avatarverse}, generating general 3D objects remains formidable due to the lack of extensive 3D data. Recent research effort has sought to leverage the guidance of pretrained text-to-image generative models~\citep{poole2022dreamfusion,lin2023magic3d,tang2023make} and showcases promising results. 

The idea of leveraging pretrained text-to-image (T2I) models for 3D generation is initially proposed by DreamFusion~\citep{poole2022dreamfusion}. A score distillation sampling (SDS) loss is enforced to optimize the 3D model such that its renderings at random viewpoints match the text-conditioned image distribution as interpreted by a powerful T2I diffusion model. DreamFusion inherits the imaginative power of 2D generative models and can yield highly creative 3D assets. To deal with the over-saturation and blurriness issues, recent works adopt stage-wise optimization strategies~\citep{mildenhall2021nerf} or propose improved 2D distillation loss~\citep{wang2023prolificdreamer}, which leads to an enhancement in photo-realism. However, the majority of current research falls short of synthesizing complex content as achieved by 2D generative models.
In addition, these works are often plagued with the ``Janus issue'', where 3D renderings that appear plausible individually show semantic and stylistic inconsistencies when examined holistically. 

\begin{figure}[t]
\begin{center}
%\framebox[4.0in]{$\;$}
\includegraphics[width=\textwidth]{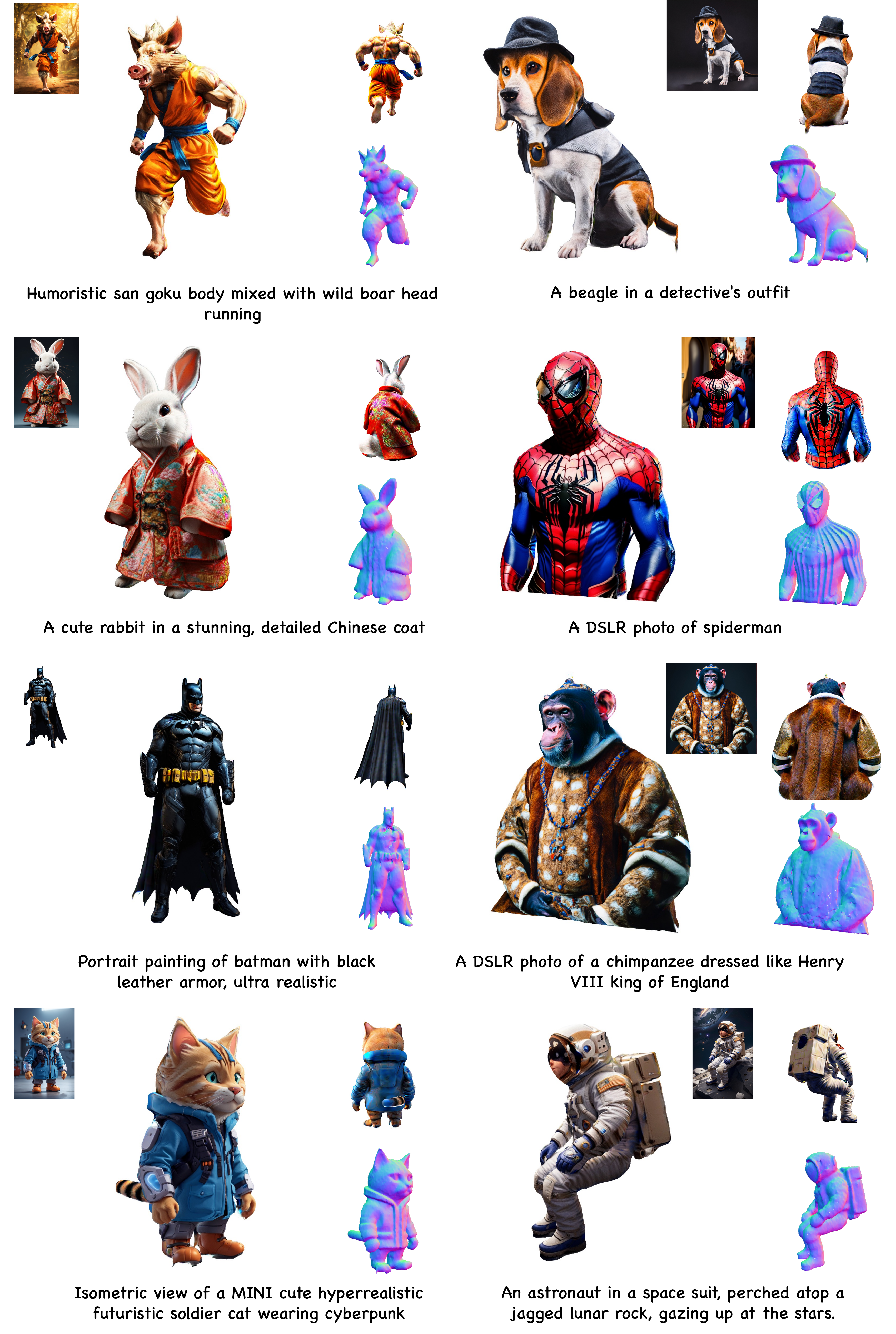}
\end{center}
\caption{By lifting 2D images to 3D, DreamCraft3D achieves 3D generation with rich details and holistic 3D consistency. Please refer to the Appendix and the demo video for more results.}
\label{eq:teaser}
\end{figure}
\clearpage

In this paper, we propose \emph{DreamCraft3D}, an approach to produce complex 3D assets while maintaining holistic 3D consistency. Our approach explores the potential of hierarchical generation. We draw inspiration from the manual artistic process: an abstract concept is first solidified into a 2D draft, followed by the sculpting of rough geometry, the refinement of the geometric details and the painting of high-fidelity textures. We adopt a similar approach, breaking down the challenging 3D generation into manageable steps. Starting with a high-quality 2D reference image generated from a text prompt, we lift it into 3D via stages of geometry sculpting and texture boosting. Contrary to prior approaches, our work highlights how careful consideration of each stage can unleash the full potential of hierarchical generation, resulting in superior-quality 3D creation.

The geometry sculpting stage aims to produce plausible and consistent 3D geometry from the 2D reference image. On top of using the SDS loss for novel views and photometric loss at the reference view, we introduce multiple strategies to promote geometric consistency. Foremost, we leverage an off-the-shelf viewpoint-conditioned image translation model, Zero-1-to-3~\citep{liu2023zero}, to model the distribution of novel views based on the reference image. Since this view-conditioned diffusion model is trained on diverse 3D data~\citep{deitke2023objaverse}, it provides a rich 3D prior that complements the 2D diffusion prior. Additionally, we find annealing the sampling timestep and progressively enlarging training views are crucial to further improve coherency. During optimization, we transition from implicit surface representation~\citep{wang2021neus} to mesh representation~\citep{shen2021deep} for coarse-to-fine geometry refinement. Through these techniques, the geometry sculpting stage produces sharp, detailed geometry while effectively suppressing most geometric artifacts.

We further propose \emph{bootstrapped score distillation} to substantially boost the texture. Existing view-conditioned diffusion models trained on limited 3D often struggle to match the fidelity of modern 2D diffusion models. Instead, we finetune the diffusion model according to multi-view renderings of the 3D instance being optimized. This personalized 3D-aware generative prior becomes instrumental in augmenting the 3D texture while ensuring view consistency. Importantly, we find that alternatively optimizing the generative prior and 3D representation leads to mutually reinforcing improvements. The diffusion model benefits from training on improved multi-view renderings, which in turn provides superior guidance for optimizing the 3D texture. In contrast to prior works~\citep{poole2022dreamfusion,wang2023prolificdreamer} that distill from a fixed target distribution, we learn from a distribution that gradually evolves according to the optimization state. Through this ``bootstrapping'', our approach captures increasingly detailed texture while keeping the view consistency.

% far greater texture diversity and fidelity than 3D priors trained on limited 3D data, which allows for generating creative objects with delicate texture details exceeding the state-of-the-art.  

As shown in Figure~\ref{eq:teaser}, our method is capable of producing creative 3D assets with intricate geometric structures and realistic textures rendered coherently in 360\degree. Compared to optimization-based approaches ~\citep{poole2022dreamfusion,lin2023magic3d}, our method offers substantially improved texture and complexity. Meanwhile, compared to image-to-3D  techniques~\citep{tang2023make,qian2023magic123}, our work excels at producing unprecedented realistic renderings in 360\degree renderings. 
% Our method produces creative objects with delicate texture details and rendering coherency, exceeding the state-of-the-art.  
% Both the qualitative and quantitative comparisons demonstrate the superiority of our approach in generation coherency and rendering realism.
These results suggest the strong potential of DreamCraft3D in enabling new creative possibilities in 3D content creation. 
The full implementation will be made publicly available.

\section{Related work}

\textbf{3D generative models} have been intensively studied to generate 3D assets without tedious manual creation. Generative adversarial networks (GANs)~\citep{chan2021pi,chan2022efficient,chan2021pi,xie2021style,zeng2022lion,skorokhodov20233d,gao2022get3d,tang2022explicitly,xie2021style,Sun_2023_CVPR,ide3d} have long been the prominent techniques in the field.  Auto-regressive models have been explored~\citep{sanghi2022clip,mittal2022autosdf,yan2022shapeformer,zhang20223dilg,yu2023pushing},  which learn the distribution of these 3D shapes conditioned on images or texts. Diffusion models~\citep{wang2023rodin,cheng2023sdfusion,li2023diffusion,nam20223d,zhang20233dshape2vecset,nichol2022point,jun2023shap,bautista2022gaudi,gupta20233dgen} have also shown significant recent success in learning probabilistic mappings from text or images to 3D shape latent. However, these methods require 3D shapes or multi-view data for training, raising challenges when generating in-the-wild 3D assets due to the scarcity of diverse 3D data~\citep{chang2015shapenet,deitke2023objaverse,wu2023omniobject3d} compared to 2D.

\textbf{3D-aware image generation} aims to render images in novel views while 
offering some level of 3D consistency. These works~\citep{sargent2023vq3d,skorokhodov20233d,xiang20233d} often rely on a pretrained monocular depth prediction model to synthesize view-consistent images. While they achieve photo-realistic renderings for categories of ImageNet, they fall short in producing results in large views. There are a few recent attempts ~\citep{watson2022novel,liu2023zero} that train view-dependent diffusion models on 3D data and demonstrate promising novel view synthesis capability for open domain. However, these are inherently 2D models and cannot ensure perfect view consistency.

\textbf{Lifting 2D to 3D} approaches improve a 3D scene representation by seeking guidance using estblished 2D text-image foundation models. Early works~\citep{jain2021dreamfields,lee2022understanding,hong2022avatarclip} utilize the pretrained CLIP~\citep{radford2021learning} model to maximize the similarity between rendered images and text prompt. DreamFusion~\citep{poole2022dreamfusion} and SJC~\citep{sjc}, on the other hand, propose to distill the score of image distribution from a pretrained diffusion model and demonstrate promising results. Recent works have sought to further enhance the texture realism via coarse-to-fine optimization~\citep{lin2023magic3d, chen2023fantasia3d}, improved distillation loss~\citep{wang2023prolificdreamer}, shape guidance~\citep{metzer2023latent} or lifting 2D image to 3D~\citep{deng2023nerdi,tang2023make,qian2023magic123,liu2023one}. 
Recently,~\citet{raj2023dreambooth3d} proposes to finetune a personalized diffusion model for 3D consistent generation. However, producing globally consistent 3D remains challenging. In this work, we meticulously design 3D priors through the whole hierarchical generation process, achieving unprecedented coherent 3D generation.

% - 3d generative models (GANs and Diffusion model)
% Diffusion-based:  SDFusion, Diffusion-SDF, 3D-LDM, 3DShape2VecSet, Rodin, Point-E and Shap-E

% GAN-based: Voxel-GAN
% normalizing flow: point flow, CLIP-Forge
% Auto-regressive: AutoSDF, shape-former, 3DILG

% \textbf{3D View-Consistency Generation}

% - 3D consistent view generation:
% 3DiM Novel View Diffusion, Zero-1-to-3, get-3d, eg3d

% However, per-scene optimization-based methods suffer from a low success rate and a long optimization time in hours to generate a high-quality 3D shape. However, they only require a pretrained CLIP or text-to-image model and do not require any 3D data.

% CLIP-guided: DreamField, PureCLIPNeRF, SinNeRF, AvatarCLIP,

% Diffusion-guided: DreamFusion, Magic3D, HiFA, ProlificDreamer, RealFusion, NeRDi, Make-It-3D, One-2-3-45, Magic-123, DreamBooth3D, SJC, LatentNeRF, Fantasia3D,  MVDreamer and SyncDreamer

\section{Preliminaries}
DreamFusion~\citep{poole2022dreamfusion} achieves text-to-3D generation by utilizing a pretrained text-to-image diffusion model $\bm{\epsilon}_\phi$ as an image prior to optimize the 3D representation parameterized by $\theta$. The image $\bm{x}=g(\theta)$, rendered at random viewpoints by a volumetric renderer, is expected to represent a sample drawn from the text-conditioned image distribution $p(\bm{x}|y)$ modeled by a pretrained diffusion model. The diffusion model $\phi$ is trained to predict the sampled noise $\bm{\epsilon}_\phi(\bm{x}_t;y,t)$ of the noisy image $\bm{x}_t$ at the noise level $t$, conditioned on the text prompt $y$. A \emph{score distillation sampling} (SDS) loss encourages the rendered images to match the distribution modeled by the diffusion model. Specifically, the SDS loss computes the gradient:
\begin{equation}
\nabla_{\theta}\mathcal{L}_\textnormal{SDS}(\phi, g(\theta))=\mathbb{E}_{t, \bm{\epsilon}}\Big[\omega(t)(\bm{\epsilon}_{\phi}(\bm{x}_{t};y,t)-\bm{\epsilon})\frac{\partial \bm{x}}{\partial \theta}\Big],
\label{eq:sds}
\end{equation}
which is the per-pixel difference between the predicted and the added noise upon the rendered image, where $\omega(t)$ is the weighting function. 

One way to improve the generation quality of a conditional diffusion model is to use the classifier-free guidance (CFG) technique to steer the sampling slightly away from the unconditional sampling, \ie,  $\epsilon_\phi(\bm{x}_t;y,t) + \epsilon_\phi(\bm{x}_t;y,t) - \epsilon_\phi(\bm{x}_t,t,\varnothing)$, where $\varnothing$ represents the ``empty'' text prompt. Typically, the SDS loss requires a large CFG guidance weight for high-quality text-to-3D generation, yet this will bring side effects like over-saturation and over-smoothing~\citep{poole2022dreamfusion}. 

Recently,~\citet{wang2023prolificdreamer} proposed a variational score distillation (VSD) loss that is friendly to standard CFG guidance strength and better resolves unnatural textures. Instead of seeking a single data point, this approach regards the solution corresponding to a text prompt as a random variable. Specifically, VSD optimizes a distribution  $q^{\mu}(\bm{x}_0|y)$ of the possible 3D representations $\mu(\theta|y)$ corresponding to the text $y$, to be closely aligned with the distribution defined by the diffusion timestep $t=0$, $p(\bm{x}_0|y)$, in terms of KL divergence:
\begin{equation}
\mathcal{L}_\textnormal{VSD} = D_\textnormal{KL}(q^{\mu}(\bm{x}_0|y) || p(\bm{x}_0|y)).
\end{equation}
\citet{wang2023prolificdreamer} further shows that this objective can be optimized by matching the score of noisy real images and that of noisy rendered images at each time $t$, so the gradient of $\mathcal{L}_\textnormal{VSD}$ is
\begin{equation}
\nabla_{\theta}\mathcal{L}_\textnormal{VSD}(\phi, g(\theta))=\mathbb{E}_{t, \bm{\epsilon}}\Big[\omega(t)(\bm{\epsilon}_{\phi}(\bm{x}_{t};y,t)-\bm{\epsilon}_\textnormal{lora}(\bm{x}_{t};y,t,c))\frac{\partial \bm{x}}{\partial \theta}\Big].
\label{eq:vsd}
\end{equation}
Here, $\bm{\epsilon}_\textnormal{lora}$ estimates the score of the rendered images using a LoRA (Low-rank adaptation)~\citep{hu2021lora} model. The obtained variational distribution yields samples with high-fidelity textures. However, this loss is applied for texture enhancement and is helpless to the coarse geometry initially learned by SDS. Moreover, both the SDS and VSD attempt to distill from a fixed target 2D distribution which only assures per-view plausibility rather than a global 3D consistency. Consequently, they suffer from the same appearance and semantic shift issue that hampers the perceived 3D quality.

\section{DreamCraft3D}
\begin{figure}[t]
\begin{center}
%\framebox[4.0in]{$\;$}
\begin{overpic}[width=\linewidth]{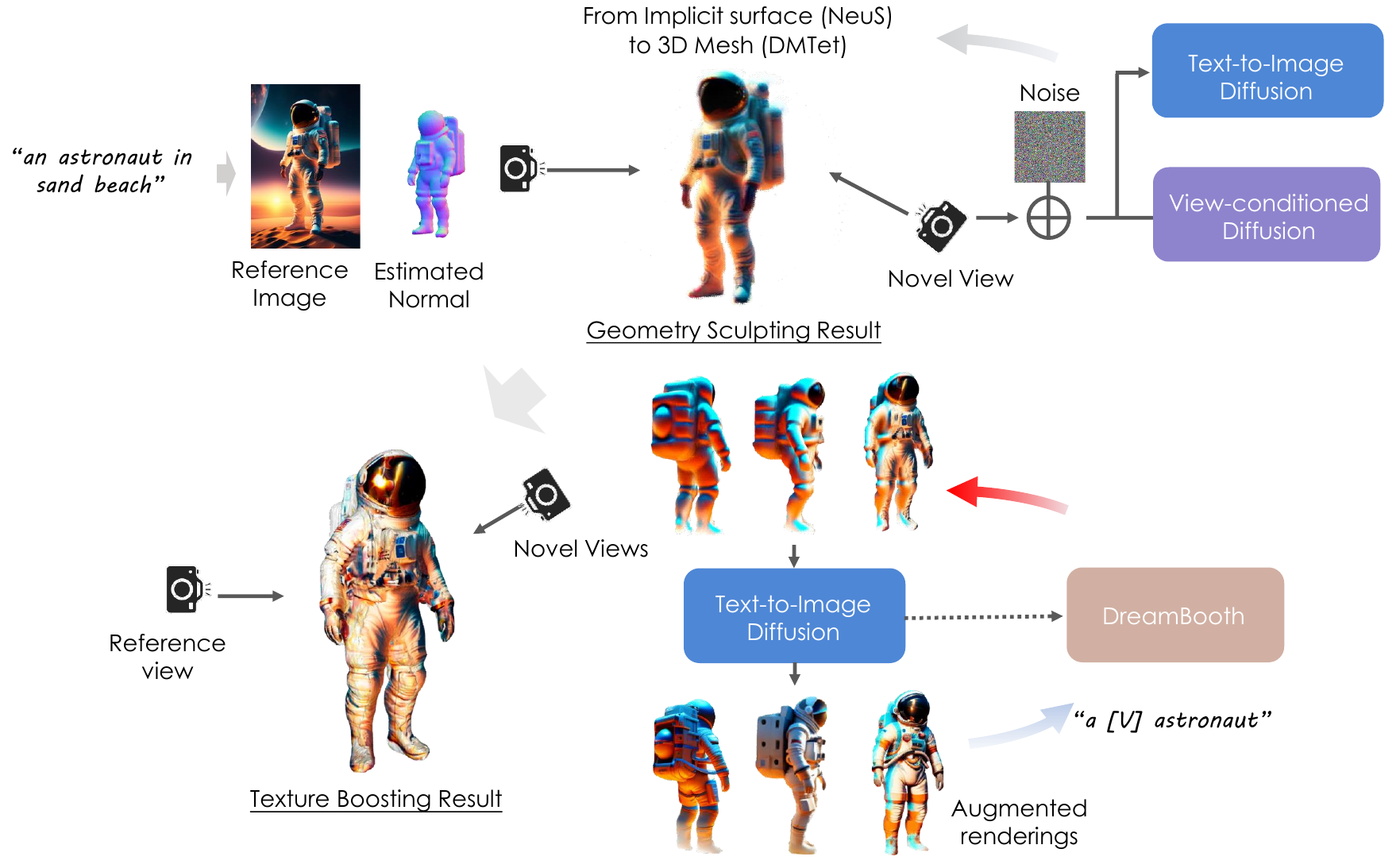}
\put(72,28){\footnotesize $\mathcal{L}_\textnormal{BSD}$}
\put(70,61){\footnotesize $\mathcal{L}_\textnormal{SDS},\mathcal{L}_\textnormal{3D-SDS}$}
\end{overpic}
\end{center}
\caption{DreamCraft3D leverages a 2D image generated from the text prompt and uses it to guide the stages of geometry sculpting and texture boosting. When sculpting the geometry, the view-conditioned diffusion model provides crucial 3D guidance to ensure geometric consistency. We then dedicately improve the texture quality by conducting a cyclic optimization. We augment the multi-view renderings and use them to finetune a diffusion model, DreamBooth, to offer multi-view consistent gradients to optimize the scene. We term the loss that distills from an evolving diffusion prior as bootstrapped distillation sampling ($\mathcal{L}_\textnormal{BSD}$ in the figure). 
}
\label{fig:diagram}
\end{figure}

We propose a hierarchical pipeline for 3D content generation as illustrated in Figure~\ref{fig:diagram}. Our method first leverages a state-of-the-art text-to-image generative model to generate a high-quality 2D image from a text prompt. In this way, we can leverage the full power of state-of-the-art 2D diffusion models to depict intricate visual semantics described in the text, retaining the creative freedom as 2D models. We then lift this image to 3D through cascaded stages of geometric sculpting and texture boosting. By decomposing the problem, we can apply specialized techniques at each stage. For geometry, we prioritize multi-view consistency and global 3D structure, allowing for some compromise on detailed textures. With the geometry fixed, we then focus solely on optimizing realistic and coherent texture, for which we jointly learn a 3D-aware diffusion prior that bootstraps the 3D optimization. In the next, we elaborate on key design considerations for the two phases. 

\subsection{Geometry sculpting}
At this stage, we aim to craft a 3D model such that it matches the appearance of the reference image $\bm{\hat{x}}$ at the same reference view while maintaining plausibility under different viewing angles. To achieve this, we encourage plausible image renderings for each randomly sampled view, recognizable by a pretrained diffusion model. This is achieved using the SDS loss  $\mathcal{L}_\textnormal{SDS}$, as defined in Equation~\ref{eq:sds}. In order to effectively utilize guidance from the reference image, we penalize the photometric difference between the rendered image and the reference via
$\mathcal{L}_{\text{rgb}} = \left\lVert \bm{\hat{m}} \odot \left(\bm{\hat{x}} - g(\theta; \hat{c})\right) \right\rVert_2$ at the reference view $\hat{c}$. The loss is computed only within the foreground region denoted by the mask $\bm{\hat{m}}$. Meanwhile, we implement the mask loss $\mathcal{L}_\textnormal{mask} = \left\Vert \bm{\hat{m}} - g_m(\theta; \hat{c})\right\Vert_2$ to encourage scene sparsity, where $g_m$ renders the silhouette. In addition, akin to~\citep{deng2023nerdi}, we fully exploit the geometry prior inferred from the reference image, and enforce the consistency with the depth and normal map computed for the reference view. The corresponding  depth and normal loss are respectively computed as:
\begin{equation}
    \mathcal{L}_\textnormal{depth}= -\frac{\text{conv}(\bm{d},\hat{\bm{d}})}{\sigma(\bm{d}) \sigma(\hat{\bm{d}})},
    \quad \mathcal{L}_\textnormal{normal}= -\frac{\bm{n} \cdot \hat{\bm{n}}}{{\Vert \bm{n} \Vert}_2 \cdot {\Vert \hat{\bm{n}} \Vert}_2},
\end{equation}
where $\text{conv}(\cdot)$ and $\sigma(\cdot)$ represent the covariance and variance operators respectively, and the depth $\hat{\bm{d}}$ and the normal $\hat{\bm{n}}$ at the reference view are computed using the off-the-shelf single-view estimator~\citep{eftekhar2021omnidata}. The depth loss adopts the form of negative Pearson correlation $\mathcal{L}_\textnormal{depth}$ to account for the scale mismatch in depth.

Despite these, maintaining consistent semantics and appearance across back-views remains a challenge. Thus, we employ additional techniques to produce coherent, detailed geometry.

\noindent\textbf{3D-aware diffusion prior.} We argue that the 3D optimization with per-view supervision alone is under-constrained. Hence, we utilize a view-conditioned diffusion model, Zero-1-to-3, which is trained on a large scale of 3D assets and offers an improved viewpoint awareness. The Zero-1-to-3 is a fine-tuned 2D diffusion model, which hallucinates the image in a relative camera pose $c$ given the reference image $\bm{\hat{x}}$. This 3D-aware model encodes richer 3D knowledge of the visual world and allows us to better extrapolate the views given a reference image. As such, we distill the probability density from this model and compute the gradient of a 3D-aware SDS loss for novel views:
\begin{equation}
\nabla_{\theta}\mathcal{L}_\textnormal{3D-SDS}(\phi, g(\theta))=\mathbb{E}_{t, \bm{\epsilon}}[\omega(t)(\bm{\epsilon}_{\phi}(\bm{x}_{t};\bm{\hat{x}},c,y,t)-\bm{\epsilon})\frac{\partial \bm{x}}{\partial \theta}].
\end{equation}
This loss effectively alleviates 3D consistency issues like Janus problem. However, the finetuning on limited categories of 3D data of inferior rendering quality impairs the diffusion model's generation capability, so the 3D-aware SDS loss alone is prone to induce deteriorated quality when lifting general images to 3D. Therefore, we employ a hybrid SDS loss, which incorporates both the 2D and 3D diffusion priors simultaneously. Formally, this hybrid SDS loss provides the gradient as:
\begin{equation}
\nabla_{\theta}\mathcal{L}_\textnormal{hybrid}(\phi, g(\theta))=\nabla_{\theta}\mathcal{L}_\textnormal{SDS}(\phi, g(\theta)) + \mu \nabla_{\theta}\mathcal{L}_\textnormal{3D-SDS}(\phi, g(\theta)),
\label{eq:hybrid}
\end{equation}
where we choose $\mu=2$ to emphasize the weight of the 3D diffusion prior. When computing $\mathcal{L}_\textnormal{SDS}$, we adopt the DeepFloyd IF base model~\citep{deep-floyd}, a diffusion model that operates at $64\times 64$ resolution pixel space and better captures coarse geometry. 

\noindent\textbf{Progressive view training.} However, directly deriving the free views in 360\degree may still result in geometric artifacts, such as extra chair legs, due to the ambiguity inherent in a single reference image. To solve this, we progressively enlarge the training views, gradually propagating the well-established geometry to 360\degree results.

\noindent\textbf{Diffusion timestep annealing.}
To align with the coarse-to-fine progression of 3D optimization, we adopt a diffusion timestep annealing strategy similar to~\cite{huang2023dreamtime}. At the start of optimization, we prioritize sampling larger diffusion timestep $t$ from the range $[0.7, 0.85]$ when computing Equation~\ref{eq:hybrid} to provide the global structure. As training proceeds, we linearly anneal the $t$ sampling range to $[0.2, 0.5]$ over hundreds of iterations. This annealing strategy allows the model to first establish a plausible global geometry in the early optimization phase before refining the structural details. 

\noindent\textbf{Detailed structural enhancement.} We initially optimize an implicit surface representation with the corresponding volume rendering as in NeuS~\citep{wang2021neus} to establish the coarse structure. Then, following~\cite{lin2023magic3d}, we use this result to initialize a textured 3D mesh representation using a deformable tetrahedral grid (DMTet) ~\citep{shen2021deep} to facilitate high-resolution details. Moreover, this representation disentangles the learning of geometry and texture. Hence, at the end of this structural enhancement, we are able to solely refine the texture and better preserve high-frequency details from the reference image.

% Building on Magic3D, we convert the coarse neural fields into a memory-efficient DMTet hybrid SDF-Mesh representation for further refinement. Unlike Fantasia3d which separately optimizes geometry and texture, our approach iteratively optimizes both in an alternating manner by rendering normal maps and RGB images sequentially. This avoids solely optimizing geometry first, for which we lack strong view supervision. With a plausible global structure already established, we forgo 3D priors during mesh optimization. Instead, following Dreamfusion, we randomly sample camera radius and FOV to enrich geometry and texture details from multiple viewpoints, improving consistency. The hybrid DMTet representation coupled with alternating geometry-texture optimization allows efficiently enhancing both global structure and fine details. By rendering from varied viewpoints, we strengthen multi-view coherence without 3D priors.

\subsection{Texture boosting via bootstrapped score sampling}
The geometry sculpting stage prioritizes the learning of coherent and detailed geometry but leaves the texture blurry. This is due to our reliance on a 2D prior model that operates at a coarse resolution, and the limited sharpness offered by the 3D-aware diffusion model. Additionally, texture issues such as over-smoothing and over-saturation arise from excessively large classifier-free guidance. 

To augment the texture realism, we use variational score distillation (VSD) loss, as detailed in Equation~\ref{eq:vsd}. We switch to the Stable Diffusion model~\citep{rombach2021highresolution} in this stage which offers high-resolution gradients. To promote realistic rendering, we exclusively optimize the mesh texture with the tetrahedral grid fixed. In this learning stage, we do not leverage the Zero-1-to-3 model as the 3D prior since it adversely impacts the texture quality. Nonetheless, the inconsistent textures may come back, resulting in bizarre 3D outcomes. 

We observe that the multi-view renderings from the last stage, despite some blurriness,  exhibit good 3D consistency. One idea is to adapt a pretrained 2D diffusion model using these rendering results, enabling the model to form a concept about the scene's surrounding views. In light of this, we finetune the diffusion model with the multi-view image renderings $\{\bm{x}\}$, using DreamBooth~\citep{ruiz2023DreamBooth}. Specifically, we incorporate the text prompts containing a unique identifier and the subject's class name (e.g., “A [V] astronaut” in Figure~\ref{fig:diagram}). During finetuning, the camera parameter of each view is introduced as an additional condition. In practice, we train the DreamBooth with ``augmented'' image renderings, $\bm{x_r} = r_{t^\prime}(\bm{x})$. We introduce Gaussian noises, in an amount specified by the diffusion timestep $t^\prime$, to the multi-view renderings, \ie, $\bm{x}_{t^\prime} = \alpha_{t^\prime} \bm{x}_0 + \sigma_{t^\prime}\bm{\epsilon}$ ($\alpha_{t^\prime}, \sigma_{t^\prime}>0$ are hyperparameters), which are restored using the diffusion model. By choosing a large $t^\prime$, these augmented images reveal high-frequency details at the cost of the fidelity to the original renderings. The DreamBooth model trained on these augmented renderings can serve as a 3D prior to guide texture refinement.

Further, we propose to alternatively optimize the 3D scene to facilitate a bootstrapped optimization (Figure~\ref{fig:diagram}). Initially, the 3D mesh yields blurry multi-view renderings. We adopt a large diffusion $t^\prime$ to augment their texture quality while introducing some 3D inconsistency. The DreamBooth model trained on these augmented renderings obtains a unified 3D concept of the scene to guide texture refinement. As the 3D mesh reveals finer textures, we reduce the diffusion noises introduced to the image renderings, so the DreamBooth model learns from more consistent renderings and better captures the image distribution faithful to evolving views.
In this cyclic process, the 3D mesh and diffusion prior mutually improve in a bootstrapped manner. Formally, we derive the 3D optimization gradient using the following bootstrapped score distillation (BSD) loss:
\begin{equation}
\nabla_{\theta}\mathcal{L}_\textnormal{BSD}(\phi, g(\theta))=\mathbb{E}_{t, \bm{\epsilon}, c}[\omega(t)(\bm{\epsilon}_\textnormal{DreamBooth}(\bm{x}_{t};y,t,r_{{t}^\prime}(\bm{x}),c)-\bm{\epsilon}_\textnormal{lora}(\bm{x}_{t};y,t,\bm{x},c))\frac{\partial \bm{x}}{\partial \theta}].
\label{eq:bsd}
\end{equation}
Contrary to prior works~\citep{poole2022dreamfusion,wang2023prolificdreamer} that distill the score function from a fixed 2D model, our BSD loss learns from an evolving model which becomes increasingly 3D consistent by drawing feedback from the ongoing crafted 3D model. In our experiments, we alternate the optimization twice, which suffices to produce consistent textures with rich details.

\section{Experiments}

\begin{figure}[t]
\begin{center}
\begin{overpic}[width=\linewidth]{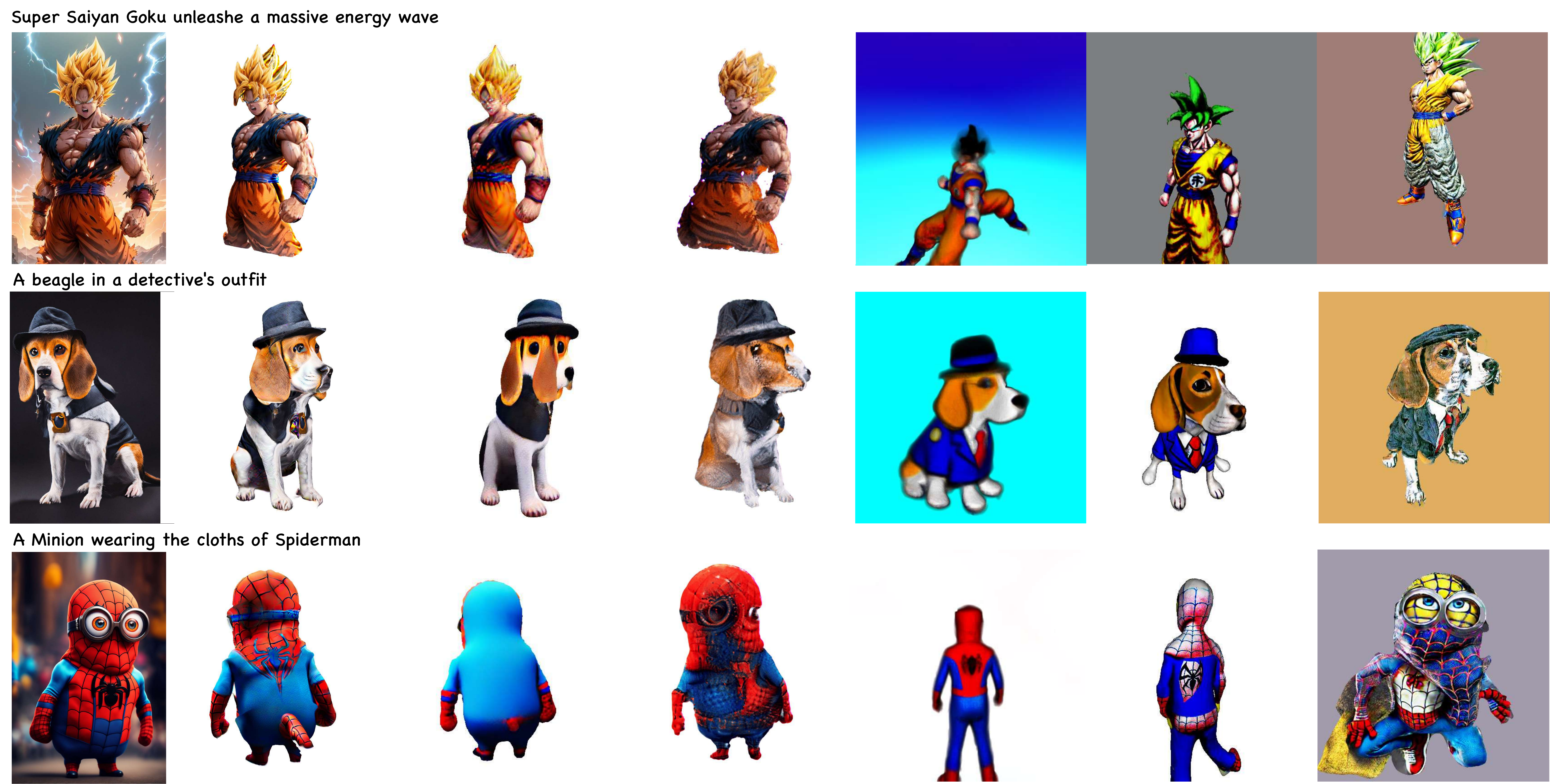}
\put(2,-2.4){\footnotesize Reference}
\put(17,-2.4){\footnotesize {Ours}}
\put(29,-2.4){\footnotesize Magic123}
\put(43,-2.4){\footnotesize Make-It-3D}
\put(57,-2.4){\footnotesize DreamFusion}
\put(73,-2.4){\footnotesize Magic3D}
\put(85,-2.4){\footnotesize ProlificDreamer}
\end{overpic}
\vspace{0.05em}
\end{center}   
\caption{Qualitative comparison with baselines. Our method generates sharper and more plausible details in both geometry and texture. Note that our method generates rich texture detail at novel views and eliminates multi-face Janus problems.}
\label{fig:qualitative}
\end{figure}

\subsection{Implementation Details}
\noindent\textbf{Architectural details.} In the geometry sculpting stage, we use Neus and textured 3D mesh representations. We employ Instant NGP~\citep{mueller2022instant}, optimizing from a 64 to a 384 resolution. For the textured mesh, we use DMTet at a 128 grid and 512 rendering resolution.

\noindent\textbf{Optimization.} 
During mesh refinement, we iteratively render a guided normal map and an RGB image, enhancing geometric detail and optimizing our texture prediction network for consistency. Considering the given plausible global geometric structure, our approach eschews the use of a 3D prior during texture optimization. We leverage random sampling of the camera radius and field-of-view (FOV) angles, aligning with Dreamfusion's methodology. This results in improved texture and geometry details via alternate rendering of normal maps and RGB images.

\vspace{-0.2em}
\begin{figure}[t]
\begin{center}
\includegraphics[width=1.0\linewidth]{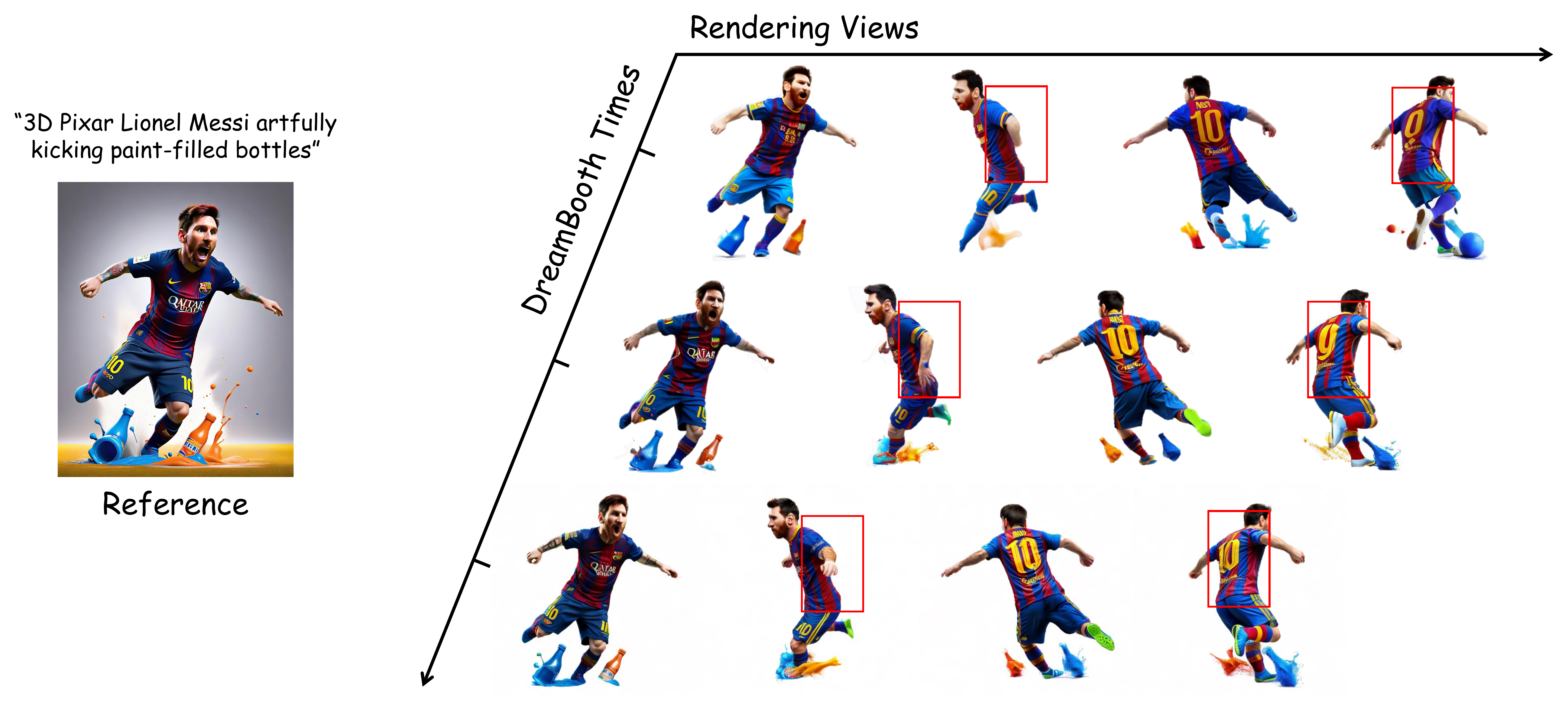}
\end{center}
\caption{Improved view consistency and texture fidelity along bootstrapping.}
\label{fig:dreambooth_time}
\end{figure}

\subsection{Comparisons with the State of the Arts}
\noindent\textbf{Baselines.} We conduct a comparative analysis of our technique against five baseline methods. The first three are text-to-3D methods: DreamFusion~\citep{poole2022dreamfusion}, Magic3D~\citep{lin2023magic3d} and ProlificDreamer~\citep{wang2023prolificdreamer}. We also compare our method against two image-to-3d methods: Make-it-3D~\citep{tang2023make} and Magic123~\citep{qian2023magic123}. For DreamFusion, Magic3D, Magic123 and ProlificDreamer, we utilize their implementations in the Threestudio library~\citep{threestudio2023} for comparison. For Make-it-3D, we use its official implementation.

\begin{wraptable}{r}{0.55\textwidth}
\vspace{-0.8em}
\centering
\small
\renewcommand{\arraystretch}{1.1}
\setlength\tabcolsep{1.5pt}
\caption{Quantitative comparison against prior 2D-to-3D lifting methods. The metrics are measured on 300 generated samples.}
\label{table:Image-to-3D}
\begin{tabular}{c|cccc}
\toprule
& CLIP $\uparrow$ & Contextual $\downarrow$ & PSNR $\uparrow$  & LPIPS $\downarrow$  \\
\midrule
Make-it-3D   & 0.872           & 1.609      & 18.937 & 0.054 \\
% \midrule
Magic123     & 0.843           & 1.628      & 22.838 & 0.053 \\
% \midrule
\textbf{DreamCraft3D} & \textbf{0.896}           & \textbf{1.579}      & \textbf{31.801} & \textbf{0.005} \\
\bottomrule
\end{tabular}
\end{wraptable}

\noindent\textbf{Datasets.} We establish a test benchmark that includes 300 images, which is a mix of real pictures and those produced by Stable Diffusion~\citep{rombach2021highresolution} and Deep Floyd. Each image in this benchmark comes with an alpha mask for the foreground, a predicted depth map, and a text prompt. For real images, the text prompts are sourced from an image caption model. We intend to make this test benchmark accessible to the public.

\noindent\textbf{Quantitative comparison.}
To generate compelling 3D content that resembles the input image and consistently conveys semantics from various perspectives, we compare our technique with established baselines using a quantitative analysis. Our evaluation employed four metrics: LPIPS~\citep{zhang2018unreasonable} and PSNR for fidelity measurement at the reference viewpoint; Contextual Distance~\citep{mechrez2018contextual} for pixel-level congruence assessment; and CLIP score~\citep{radford2021learning} to estimate semantic coherence. Results depicted in Table~\ref{table:Image-to-3D} indicate that our approach significantly surpasses the baselines in maintaining both texture consistency and fidelity.

\begin{wrapfigure}{r}{0.4\linewidth}
\centering
%\framebox[4.0in]{$\;$}
\includegraphics[width=1.0\linewidth]{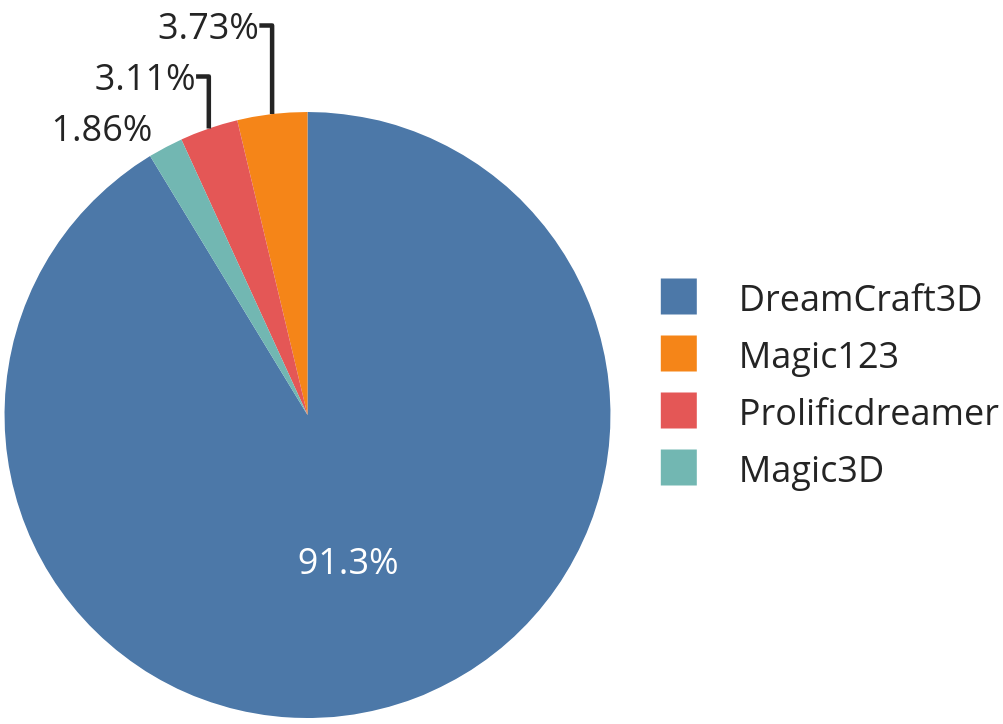}
\vspace{0.01em}
\caption{User study.}
\label{fig:user_study}
\end{wrapfigure}

\noindent\textbf{User study.} To substantiate the robustness and quality of our proposed model, we executed a user study employing 15 distinct pairs of prompts and images. Each participant was provided with four free-view rendering video alongside their corresponding textual input, and asked to choose their top preferred 3D model. The study gathered 480 responses from a total of 32 participants, the analysis of which is depicted in Figure~\ref{fig:user_study}. On an average basis, our model was favored by 92\% of users over alternative models, outperforming the baselines by a large margin. This result provides compelling evidence of the resilience and superior quality inherent to our proposed method.

% \begin{table}[t]
% % \captionsetup{font=normal}
% \setlength{\tabcolsep}{10pt}
% \footnotesize
% \begin{center}
% \begin{tabularx}{1.0\linewidth}{X cccc}
% \toprule
% Model & PSNR$\uparrow$ & LPIPS$\downarrow$ & CLIP$\uparrow$ \\
% \midrule            
% Dreamfusion                       & N/A & $14.75\pm 0.81$ & $31.31\pm3.34$ \\
% % \; - larger elevation range        & - & $32.64$ & $14.26\pm0.72$  & \\
% \; Make-it-3D     & $33.41$ & $12.76\pm0.70$ & $30.60\pm3.14$\\
% \; Magic123                       & $32.57$ & $13.72\pm 0.91$ & $31.40\pm3.05$ \\
% \; Ours   & $32.06$ & $13.68\pm0.41$ & $31.31\pm3.12$\\
% \bottomrule
% \end{tabularx}
% \vspace{-1.0em}\caption{Quantitative evaluation on image synthesis quality. All models are sampled using DDIM sampler.}\vspace{-2.5em}
% \label{tab:synthesis_quality}
% \end{center}
% \end{table}

% \begin{table}[t]
% \centering
% \small
% \renewcommand{\arraystretch}{1.1}
% \caption{Quantitative comparison against prior 2D-to-3D lifting methods. The metrics are measured on 300 generated samples.}
% \label{table:Image-to-3D}
% \begin{tabular}{c|cccc}
% \toprule
% & CLIP $\uparrow$ & Contextual $\downarrow$ & PSNR $\uparrow$  & LPIPS $\downarrow$  \\
% \midrule
% Make-it-3D   & 0.872           & 1.609      & 18.937 & 0.054 \\
% % \midrule
% Magic123     & 0.843           & 1.628      & 22.838 & 0.053 \\
% % \midrule
% \textbf{DreamCraft3D} & \textbf{0.896}           & \textbf{1.579}      & \textbf{31.801} & \textbf{0.005} \\
% \bottomrule
% \end{tabular}
% \end{table}

\noindent\textbf{Qualitative comparison.} % • our lifting 2D to 3D pipeline yield better imagination diversity
Figure~\ref{fig:qualitative} compares our method with the baselines. All the text-to-3D methods suffer from multi-view consistency issues. While ProlificDreamer offers realistic textures, it fails to form a plausible 3D object. Image-to-3D methods like Make-it-3D create quality frontal views but struggle with geometry. Magic123, enhanced by Zero1-to-3, fares better in geometry regularization, but both generate overly smoothed textures and geometric details. In contrast, our Bootstrapped Score Distillation method improves imagination diversity while maintaining semantic consistency.

\begin{figure*}[t]
    \center
    \small
    \setlength\tabcolsep{1pt}
    \renewcommand{\arraystretch}{1}
    {
    \begin{tabular}{@{}ccccc@{}}
        \includegraphics[width=0.32\textwidth]{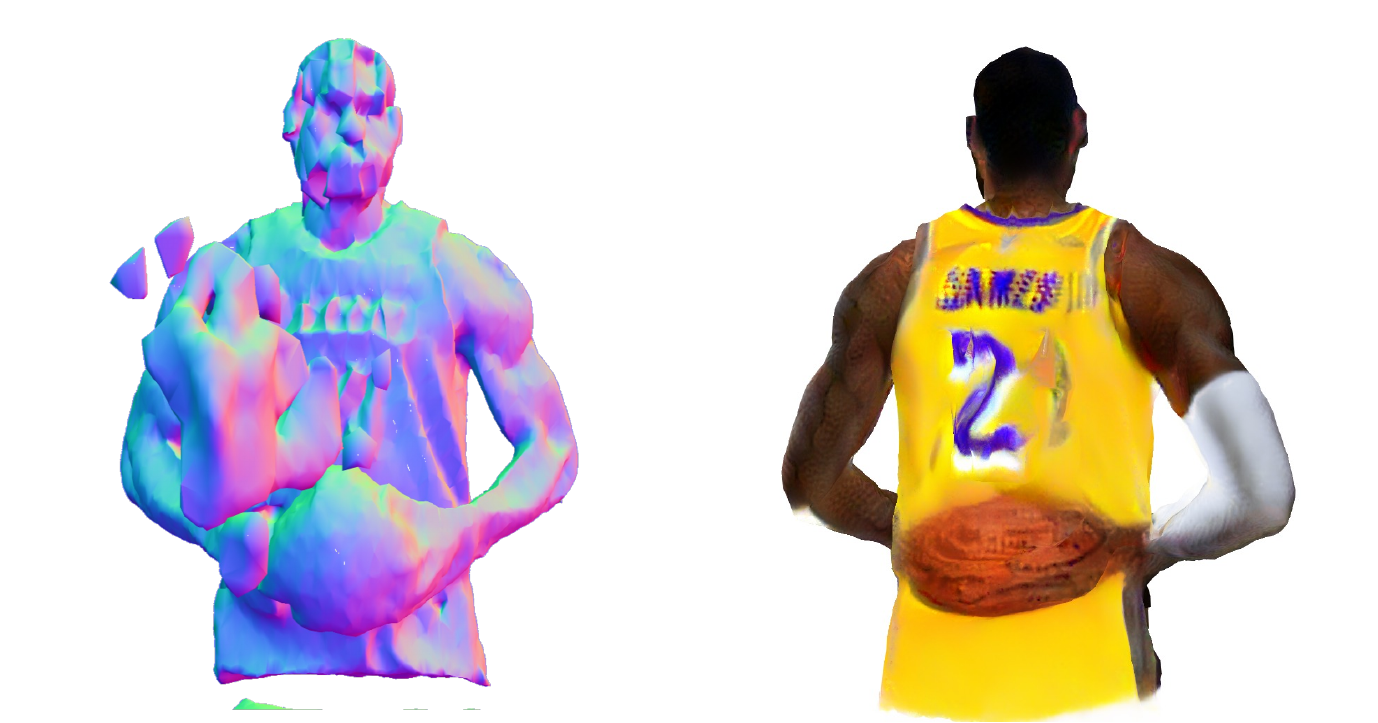} &
         \includegraphics[width=0.16\textwidth]{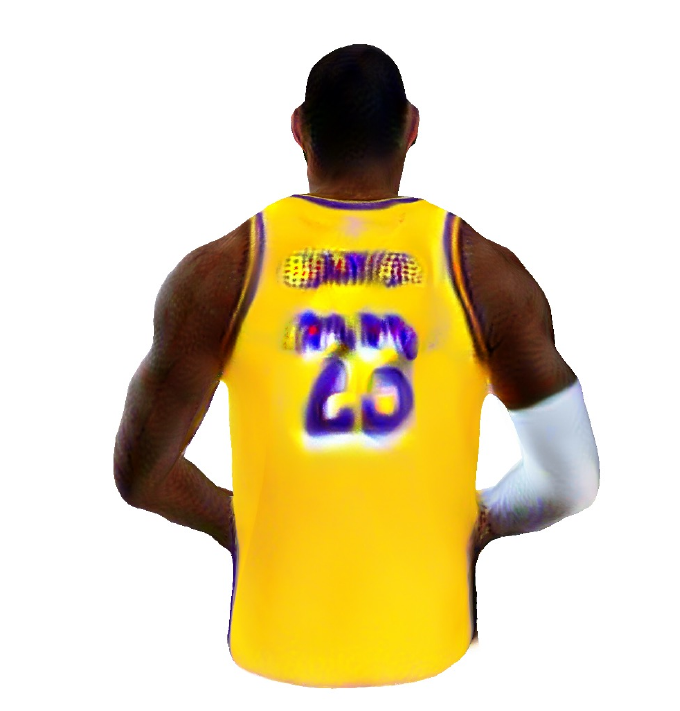} & \includegraphics[width=0.16\textwidth]{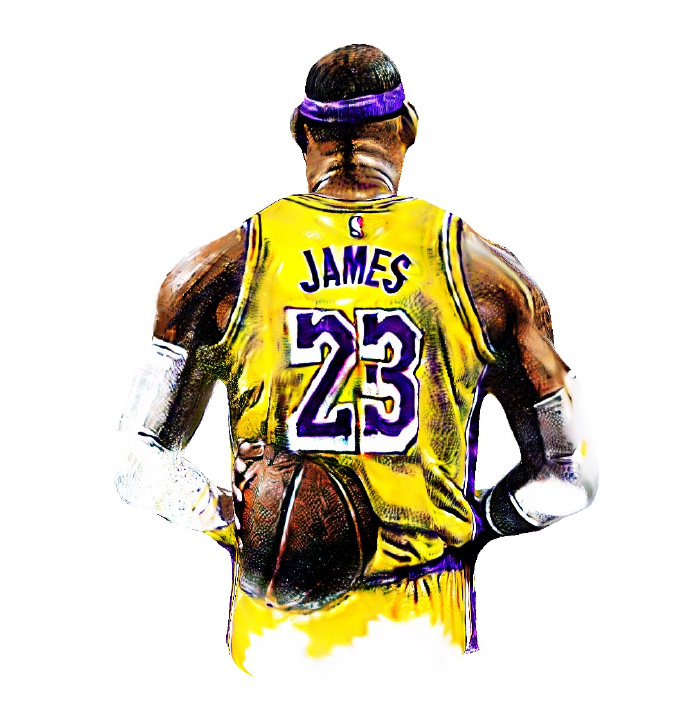} & 
         \includegraphics[width=0.16\textwidth]{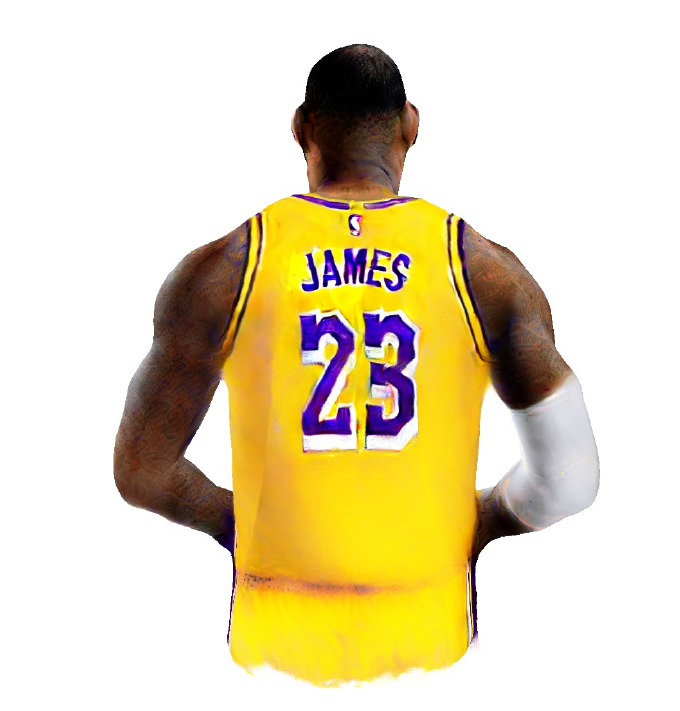} & 
         \includegraphics[width=0.16\textwidth]{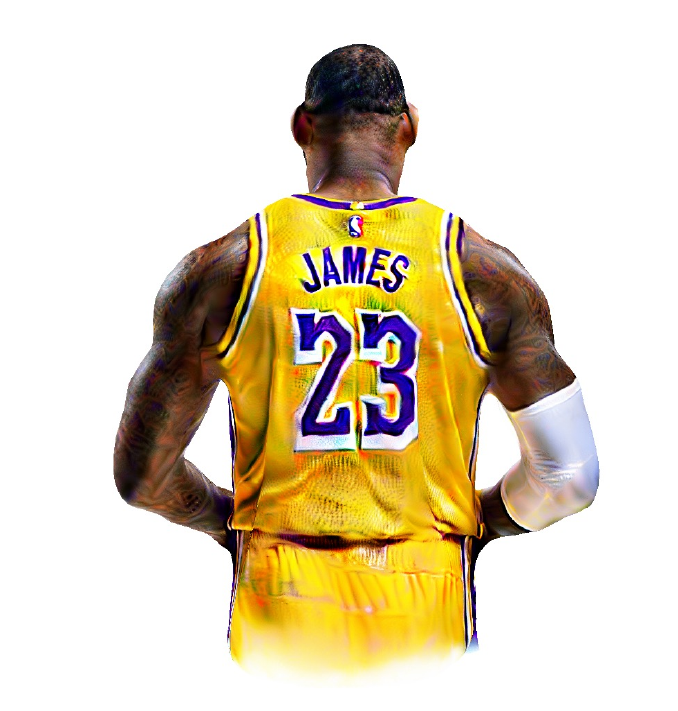} \\
         (a) w/o Zero1-to-3 & (b) SDS & (c) VSD & (d) 1-round BSD & (e) 2-round BSD
    \end{tabular}
    }
    \caption{Ablation study of the effectiveness of 3D prior and our proposed BSD (Bootstrapped Score Distillation). (a) geometry sculpting stage without 3D prior. (b) texture optimization with SDS loss. (c) VSD loss produces richer texture detail while suffering from texture inconsistency. (d) BSD improves the texture consistency with one round DreamBooth. (e) Two-round BSD adds more details to the generated result.}
    \label{fig:ablation}
\end{figure*}

\begin{figure}[t]
\begin{center}
%\framebox[4.0in]{$\;$}
\includegraphics[width=0.8\linewidth]{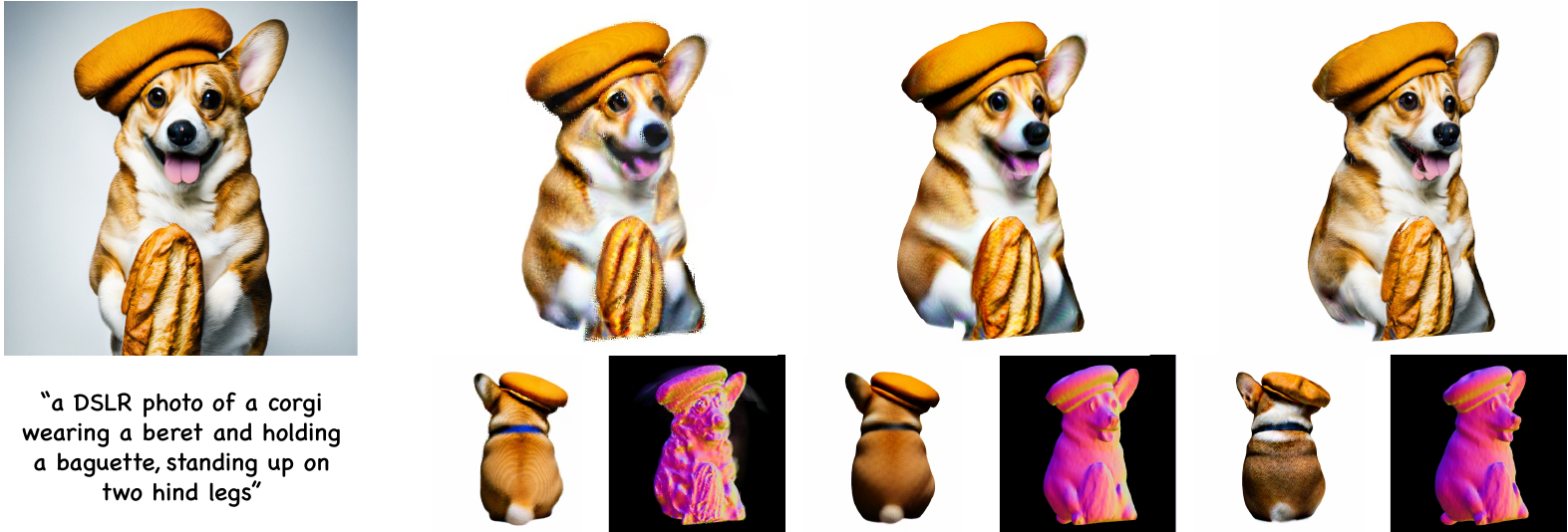}
\end{center}
\caption{Continual improvement of geometry and texture quality through multiple stages.}
\label{fig:stage_visualization}
\end{figure}

% \begin{figure}[t]
% \begin{center}
% %\framebox[4.0in]{$\;$}
% \includegraphics[width=1.0\linewidth]{Images/application/diversity.pdf}
% \end{center}
% \caption{DreamCraft3D skillfully generates an assortment of visually compelling 3D models when provided with a textual description.\bo{to update}}
% \label{fig:diversity}
% \end{figure}
\subsection{Analysis}
\noindent\textbf{The effect of 3D prior.} In our paper we claim that the guidance offered by a 3D prior enhances the generation of globally plausible geometry. To ascertain its impact, an ablation study is conducted, where the 3D prior is deactivated. Figure~\ref{fig:ablation} demonstrates that, in the absence of the 3D prior, the resultant character tends to exhibit the multifaceted Janus issue and suffers from irregular geometry. This observation underlines the significance of a viewpoint-aware 3D prior in regulating a globally consistent shape.

\noindent\textbf{The effect of BSD.} Figure~\ref{fig:ablation} also presents an ablation study encompassing three texture optimization techniques: (1) BSD, (2) VSD, and (3) Score Distillation Sampling  (SDS) with the traditional stable diffusion. The application of SDS has been observed to generate novel-view textures that are excessively smooth and over-saturated. In contrast, while VSD using standard stable diffusion can produce realistic textures, it yields a notably high inconsistency. In contrast, our proposed approach successfully generates textures that strike a balance between realism and consistency.

\noindent\textbf{Visualization of multiple stages.} Figure~\ref{fig:stage_visualization} provides the visualization of the intermediate rendering results for each stage in our hierarchical pipeline. In the geometry sculpting stage, we convert Neus to DMTet to improve high-resolution geometry details. However, the improvement in texture is negligible.  On the contrary, in the texture stage, we significantly improve the texture quality with our proposed BSD.

\noindent\textbf{DreamBooth times.} Figure~\ref{fig:dreambooth_time} illustrates multi-view datasets for DreamBooth. The initial stage involves the introduction of substantial noise into each image to amplify detail richness, leading to inconsistent denoised images. However, as the textured mesh undergoes optimization, the produced renderings evolve towards increased consistency and photorealism, thereby enhancing the quality of the input dataset tailored for DreamBooth.

\section{Conclusion}
We have presented DreamCraft3D, an innovative approach that advances the field of complex 3D asset generation. This work introduces a meticulous geometry sculpting phase for producing plausible and coherent 3D geometries and a novel Bootstrapped Score Distillation strategy. The latter, by distilling from an optimizing 3D-aware diffusion prior and adapting to multi-view renderings of the instance being optimized, significantly improves texture quality and consistency. DreamCraft3D produces high-fidelity 3D assets with compelling texture details and multi-view consistency. We believe this work represents an important step towards democratizing 3D content creation and shows great promise in future applications.

\bibliography{iclr2024_conference}
\bibliographystyle{iclr2024_conference}

\appendix
% \section{Appendix}

\section{Appendix}
\subsection{Implementation details}

\noindent\textbf{Algorithms for Bootstrapped Score Distillation.} We provide a summarized algorithm of bootstrapped score distillation in Algorithm \ref{alg:bsd}. The ``Bootstrapped Score Distillation" algorithm starts by initializing $n$ ($n=1$ in our case) meshes and a pretrained text-to-image diffusion model $\epsilon_{\text{DreamBooth}}$ parameterized by $\phi$. The algorithm then enters an iterative loop: in each iteration, it renders the mesh to obtain multi-view images $x$, augments these images with Gaussian noises to form $x_{t'} = \alpha_{t'} x + \sigma_{t'}\epsilon$, and fine-tunes the pretrained diffusion model $ \epsilon_{\text{DreamBooth}}$ based on these augmented image renderings. Within each iteration, there's an inner loop running for $n$ steps, where a random mesh and camera pose are sampled, and a 2D image is rendered from the chosen pose. Then, updates are performed on $\theta$ and $\phi$ using gradients calculated from the difference between the pretrained score function and the predicted score function, and from the $\mathcal{L}_2$ norm between the predicted score and real noise, respectively. These iterations continue until convergence, and the final refined mesh(s) are returned.

\begin{algorithm}[tb]
\caption{Bootstrapped Score Distillation}
\label{alg:bsd}
\textbf{Input: }{Number of particles $n$ $(\ge 1)$. Pretrained text-to-image diffusion model 
$\mathbf{\epsilon}_{\text{DreamBooth}}$. Learning rate $\eta_{1}$ and $\eta_{2}$ for 3D structures and diffusion model parameters, respectively. A prompt $y$. Number of images $m$ and Camera poses $\{c_{r}^{(i)}\}_{i=1}^{m}$ for the multi-view datasets.}

\begin{algorithmic}[1]
\STATE {\bfseries initialize} $n$ meshes $\{\theta^{(i)}\}_{i=1}^{n}$, a noise prediction model $\mathbf{\epsilon}_{\phi}$ parameterized by $\phi$.
\WHILE{not converged}
\STATE Render the mesh to get multi-view images $\bm{x} = g(\theta, c_{r})$. 
\STATE Augment image renderings with Gaussian noises: $\bm{x}_{t^\prime} = \alpha_{t^\prime} \bm{x} + \sigma_{t^\prime}\bm{\epsilon}$.
\STATE Finetune $\epsilon_{\textnormal{DreamBooth}}$ on augmented image renderings $\bm{x_r} = r_{t^\prime}(\bm{x})$.
\FOR{i in $T$ steps}
\STATE Randomly sample $\theta\sim \{\theta^{(i)}\}_{i=1}^{n}$ and a camera pose $c$.
\STATE Render the 3D structure $\theta$ at pose $c$ to get a 2D image $\vx_0=\vg(\theta,c)$.
\STATE $\theta \leftarrow \theta - \eta_{1} \mathbb{E}_{t,\boldsymbol{\epsilon},c}\left[
    \omega(t)
    \left( \epsilon_{\text{DreamBooth}}(\mathbf{x}_t,t,y) - \epsilon_{\phi}(\mathbf{x}_t,t,c,y)  \right)
    \frac{\partial g(\theta,c)}{\partial \theta}
\right]$
\STATE $\phi \leftarrow \phi - \eta_{2}\nabla_{\phi} \mathbb{E}_{t,\mathbf{\epsilon}}||\boldsymbol{\epsilon}_{\phi}(\mathbf{x}_t,t,c,y)-\epsilon||_2^2\mbox{.}$
\ENDFOR

\ENDWHILE
\STATE {\bfseries return} 
\end{algorithmic}
\end{algorithm}

% \begin{algorithm}
% \caption{Bootstrap Score Distillation (BSD)}
% \label{alg:bsd}
% \textbf{Input: } 3D scene parameter $\theta$, Initial camera parameters $c$, Text prompts $y$, Diffusion timestep $t^\prime$, Render function $g(\cdot)$, Pretrained Diffusion model $\epsilon_{\textnormal{DreamBooth}}$, Noise prediction model $\epsilon_{\textnormal{lora}}$. 

% \begin{algorithmic}[1]
% \STATE Initialize the 3D mesh.
% \WHILE{not converged}
%     \STATE Render the 3D structure to get multi-view images $\bm{x} = g(\theta)$. 
%     \STATE Augment image renderings with Gaussian noises: $\bm{x}_{t^\prime} = \alpha_{t^\prime} \bm{x}_0 + \sigma_{t^\prime}\bm{\epsilon}$.
%     \STATE Finetune $\epsilon_{\textnormal{DreamBooth}}$ on augmented image renderings $\bm{x_r} = r_{t^\prime}(\bm{x})$.
%     \STATE Optimize the 3D scene using the BSD loss:
%     \STATE $\nabla_{\theta}\mathcal{L}_\textnormal{BSD}(\phi, g(\theta))=\mathbb{E}_{t, \bm{\epsilon}, c}[\omega(t)(\bm{\epsilon}_\textnormal{DreamBooth}(\bm{x}_{t};y,t,r_{{t}^\prime}(\bm{x}),c)-\bm{\epsilon}_\textnormal{lora}(\bm{x}_{t};y,t,\bm{x},c))\frac{\partial \bm{x}}{\partial \theta}]$.
%     \STATE Reduce the diffusion noises introduced to image renderings.
% \ENDWHILE
% \STATE {\bfseries return} Refined 3D structure parameters $\theta$ and finetuned model $\epsilon_{\textnormal{DreamBooth}}$.
% \end{algorithmic}
% \end{algorithm}

\noindent\textbf{Structure-aware latent regularization.} To maintain the high-quality output produced by BSD while reducing noise and inconsistencies, we further incorporate a control net-guided inpainting diffusion model that regularizes the generated textures. Specifically, for a rendered image x from an arbitrary viewpoint, the visible section under the reference view is initially computed. This invariant portion during the generation process allows our inpainting model to fill in the remaining segments. As these remaining parts adhere to geometric constraints, we integrate geometric normal information through a control net. Ultimately, this method permits us to enforce view consistency and generate realistic results using a control-net guided inpainting diffusion model. To preserve the high-quality generation output, we avoid utilizing this image directly as a loss against the rendered image. Instead, we subtly introduce it by constraining the norm of the latent variables:

\begin{equation}
\mathcal{L}_\textnormal{reg}(\phi, g(\theta))=\Sigma (\|E(x)\|_2 - \|E(x_\textnormal{reg})\|_2)^2.
\end{equation}

\noindent\textbf{Architectural details.} In the Neus approach~\citep{wang2021neus}, we employ a single-layer Multi-Layer Perceptron (MLP) with 32 hidden units to simultaneously predict RGB color, volume density, and normal. The inputs to this MLP are the concatenated feature vectors derived from multi-resolution hash encoding sampled with trilinear interpolation. To sparsify the Instant NGP representation, we implement density-based pruning every 10 iterations within an octree structure, as suggested by Magic3D~\citep{lin2023magic3d}. In our experiments, we use a bounding sphere with a radius of 2. For the density prediction, we utilise the softplus activation function and, following the approach of \citealt{poole2022dreamfusion}, include an initial spatial density bias in order to encourage optimization in favor of the object-centric neural field.

\noindent\textbf{Camera and light augmentations.} We follow Magic3D to add random augmentations to the camera and light sampling for rendering the shaded images. Differently, we sample the point light location such that the angular distance from the random camera center location (w.r.t. the origin) is sampled from $\phi_\textnormal{cam} \sim \mathcal{U}(0, \pi/3)$ with a random point light distance $r_\textnormal{cam} \sim (7.5, 10)$, and (b) we freeze the material augmentation unlike Dreamfusion and Magic3D, as we found it is bad for training convergence (c) In the coarse neus stage, we propose a fixed-random mixed camera pose strategy. Specifically, following the common practice of the current text-to-3D methods, random camera view sampling benefits scene optimization. However, Zero1-to-3 needs fixed camera intrinsic parameters. Therefore, we let half of the GPUs sample the camera distance from $\mathcal{U}(3.2,3.5)$, and the Field-of-View from $\mathcal{U}(10, 20)$, while the left GPUs are fixed to the default camera intrinsic.

\noindent\textbf{Time annealing.} At the beginning of the geometry sculpting stage, We utilize a simple two-stage annealing of time step t in the score distillation objective. For the first several iterations we sample time steps $t \sim \mathcal{U}(0.7, 0.85)$ and then anneal into $t \sim \mathcal{U}(0.2, 0.50)$. We refer the readers to ProlificDreamer~\citep{wang2023prolificdreamer}. For the left iterations, we fix time steps to $\mathcal{U}(0.2, 0.50)$. We also utilize a simple two-stage time annealing for the multi-view dataset generation, that is, for the first updating step, we select a time step $t=0.5$ for all rendered images and then anneal it into $t=0.1$ along the later updating steps.
\subsection{Additional experiments}

\begin{figure}[t]
\begin{center}
%\framebox[4.0in]{$\;$}
\includegraphics[width=1.0\linewidth]{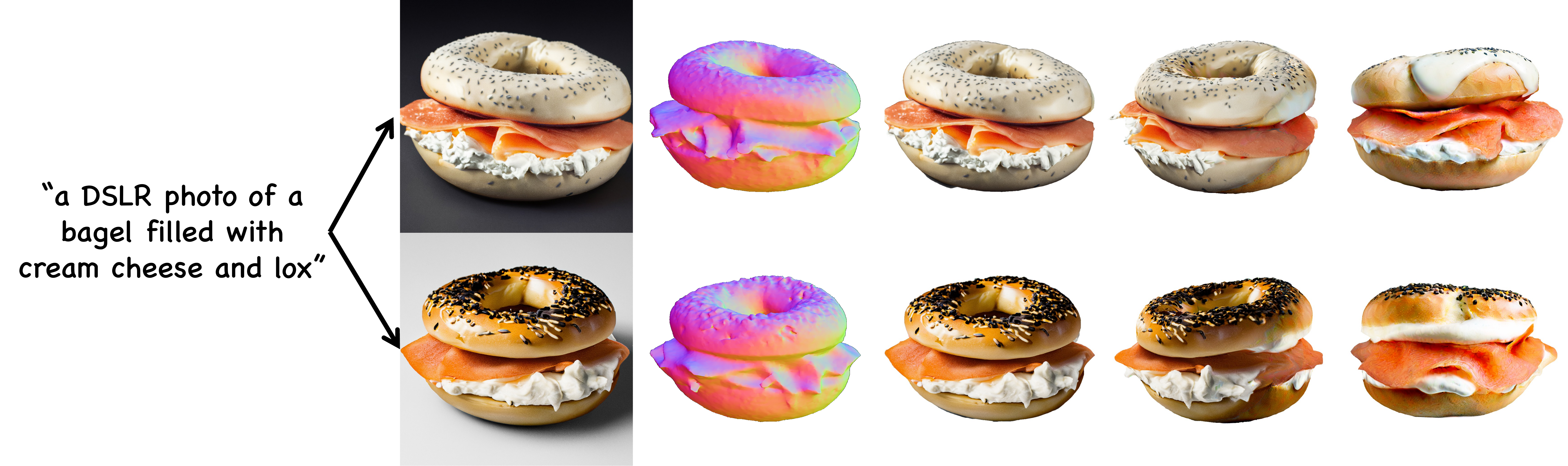}
\end{center}
\caption{DreamCraft3D skillfully generates an assortment of visually compelling 3D models when provided with a textual description.}
\label{fig:diversity}
\end{figure}

\begin{figure}[t]
\begin{center}
\includegraphics[width=1.0\linewidth]{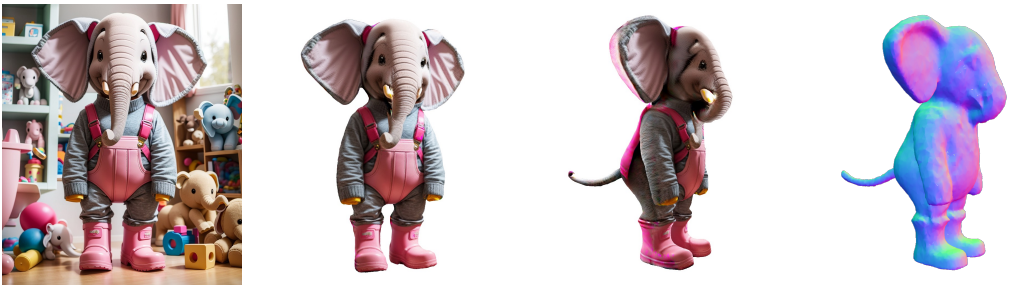}
\end{center}
\caption{Failure case. Our method learns incorrect geometry for elephant nose.}
\label{fig:limitation}
\end{figure}

\noindent\textbf{Diversity.} Prior studies frequently yield models of limited diversity with disproportionately smooth textures. Our approach to superior text-to-3D generation initially translates the text prompt into a reference image via 2D diffusion before implementing our proprietary image-based 3D creation methodology. Figure~\ref{fig:diversity} demonstrates the proficiency of our method in generating an array of diverse models from a single text prompt, all characterized by their remarkable quality.

% \begin{wrapfigure}{r}{0.45\linewidth}
% \centering
% %\framebox[4.0in]{$\;$}
% \includegraphics[width=1.0\linewidth]{Images/suppl/user_study.png}
% \vspace{0.01em}
% \caption{User study.}
% \label{fig:user_study}
% \end{wrapfigure}

% \noindent\textbf{User study.} To substantiate the robustness and quality of our proposed model, we executed a user study employing 15 distinct pairs of prompts and images. Each participant was provided with four free-view rendering video alongside their corresponding textual input, and asked to choose their top preferred 3D model. The study gathered 480 responses from a total of 32 participants, the analysis of which is depicted in Figure~\ref{fig:user_study}. On an average basis, our model was favored by 92\% of users over alternative models, outperforming the baselines by a large margin. We argue that this result provides compelling evidence of the resilience and superior quality inherent to our proposed method. 

\subsection{Limitations}
Our approach occasionally incorporates frontal-view geometric details into texture, as depicted in Figure~\ref{fig:limitation}, due to depth ambiguity and inaccuracies in the depth prior. Furthermore, we do not expressly segregate material and lighting from the 2D reference image, an aspect deferred for future exploration.

\subsection{Additional qualitative results}
\begin{figure}[t]
\begin{center}
\includegraphics[width=1.02\linewidth]{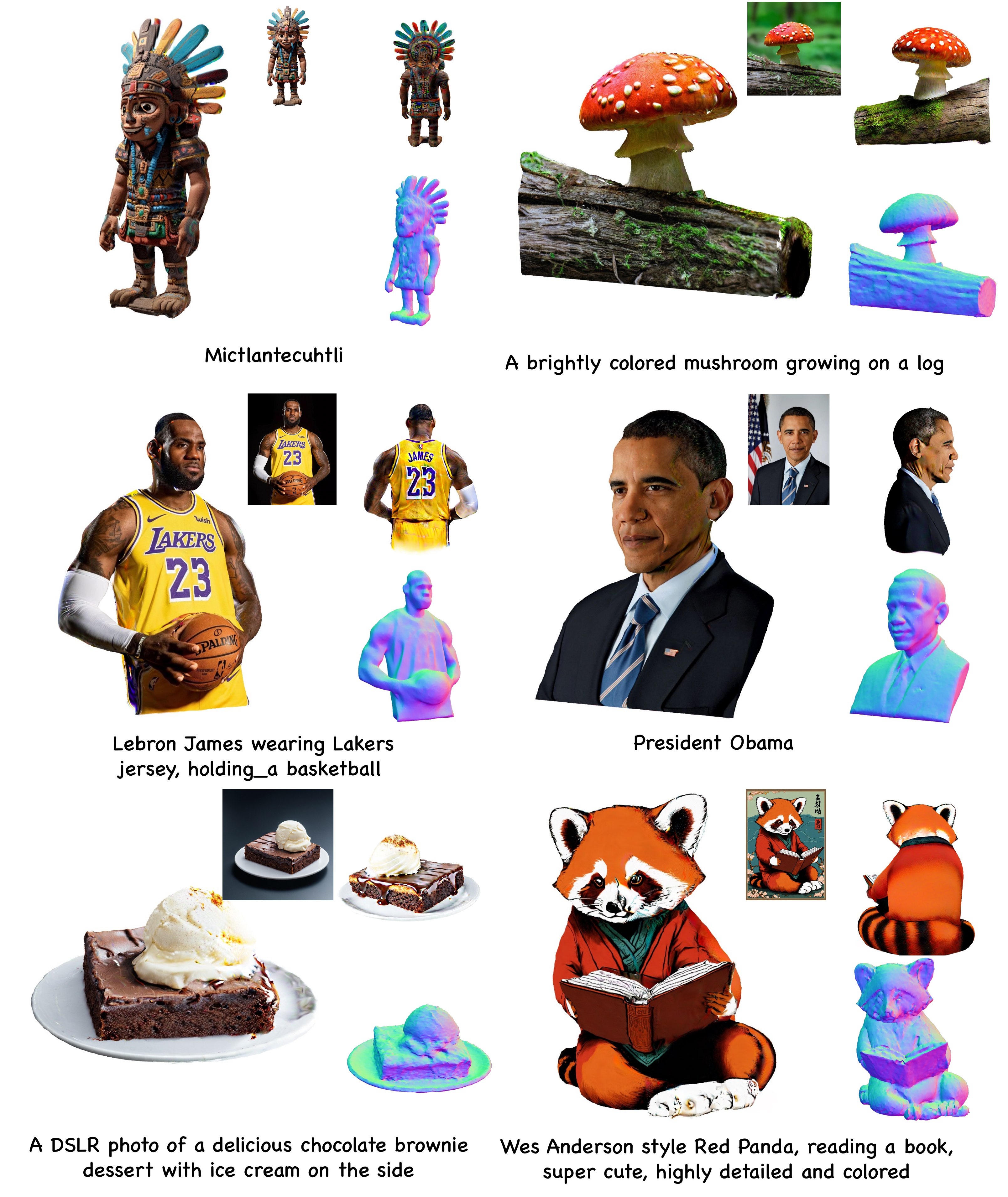}
\end{center}
\caption{Additional results of DreamCraft3D.}
\label{fig:suppl_results_0}
\end{figure}

\begin{figure}[t]
\begin{center}
\includegraphics[width=1.02\linewidth]{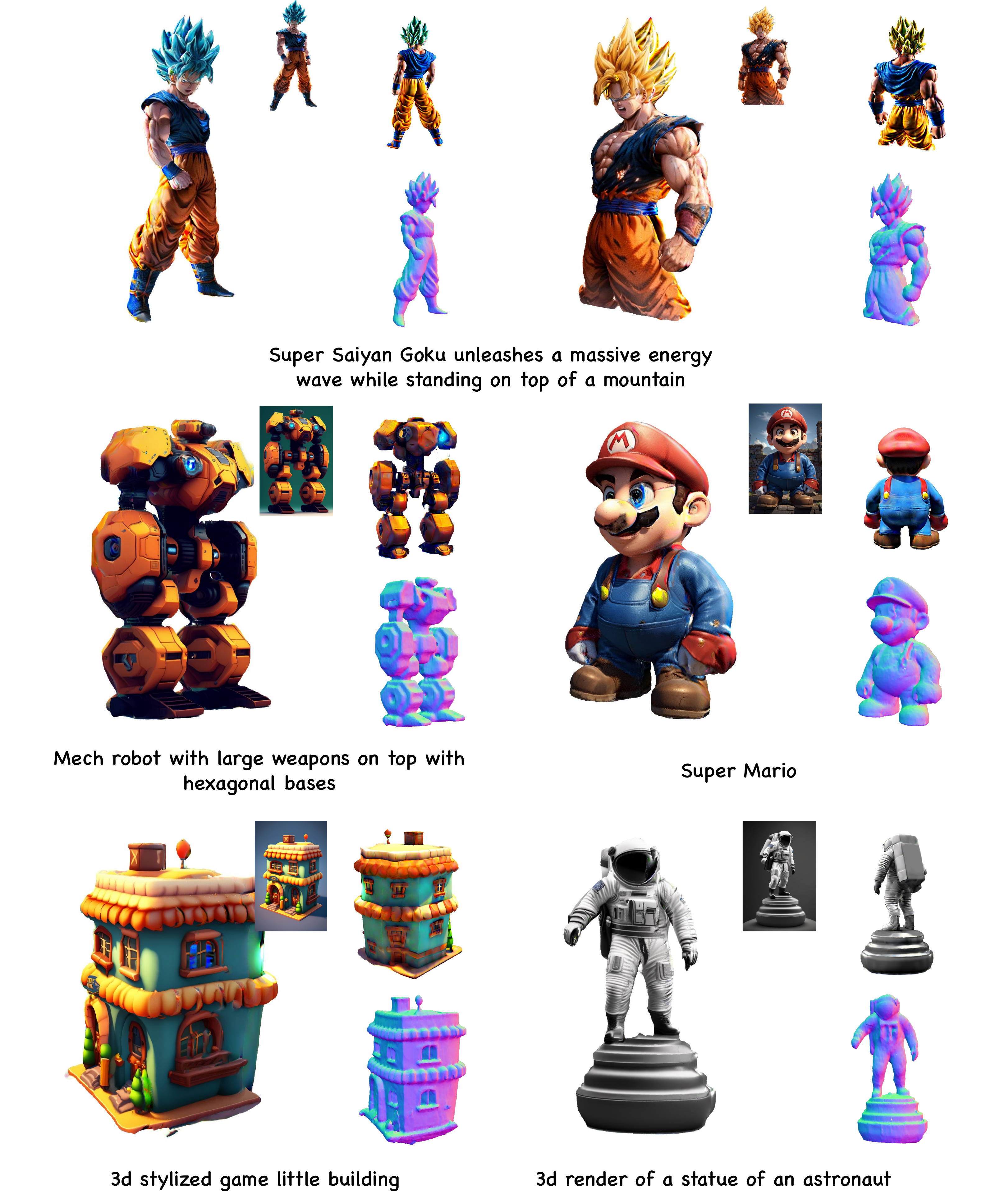}
\end{center}
\caption{Additional results of DreamCraft3D.}
\label{fig:suppl_results_1}
\end{figure}

\begin{figure}[t]
\begin{center}
\includegraphics[width=1.02\linewidth]{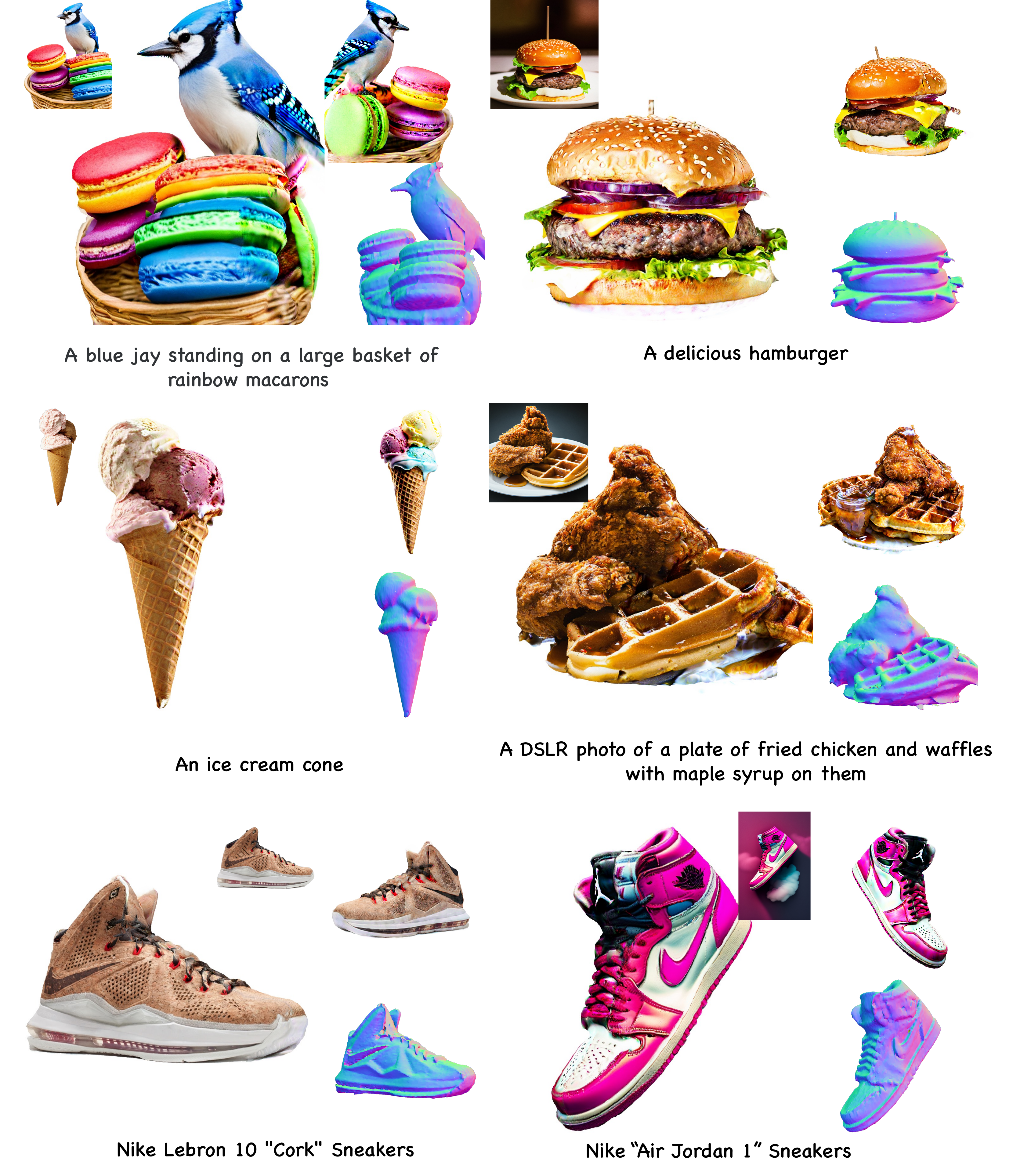}
\end{center}
\caption{Additional results of DreamCraft3D.}
\label{fig:suppl_results_2}
\end{figure}

\begin{figure}[t]
\begin{center}
\includegraphics[width=1.02\linewidth]{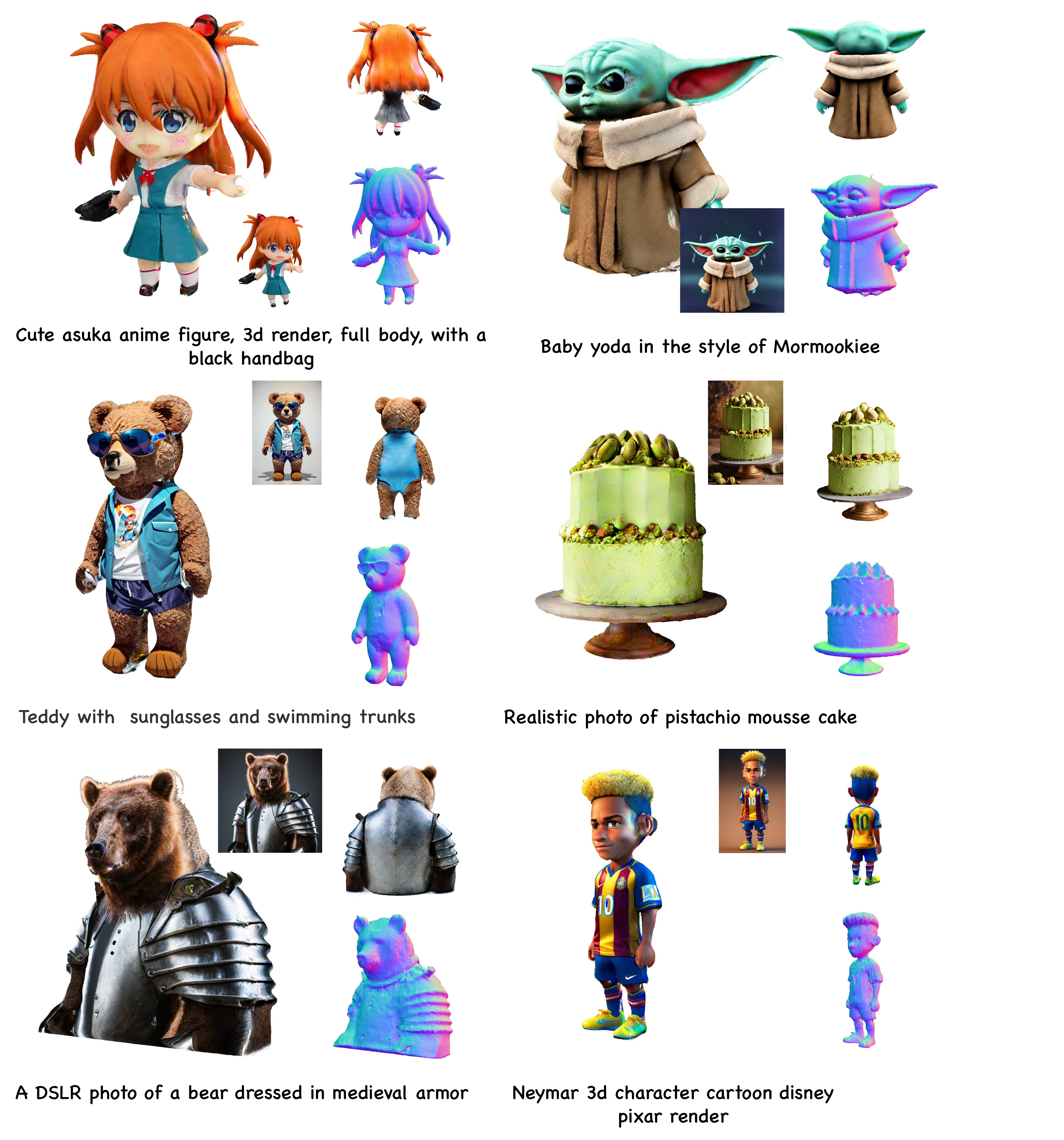}
\end{center}
\caption{Additional results of DreamCraft3D.}
\label{fig:suppl_results_3}
\end{figure}

Figure~\ref{fig:suppl_results_0}$-$Figure~\ref{fig:suppl_results_3} provides more results produced by DreamCraft3D. Our method is able to produce photo-realistic 3D assets with compelling textural details. Moreover, our method shows significantly improved 3D consistency. Please find video results in the supplementary video.

\end{document}